\documentclass[10pt,twocolumn,letterpaper]{article}

\usepackage[pagenumbers]{cvpr} 

\usepackage[dvipsnames]{xcolor}
\usepackage{pifont}
\newcommand{\cmark}{\ding{51}} 
\newcommand{\xmark}{\ding{55}} 

\newcommand{\firsttxt}[1]{\colorbox{red!15}{#1}}
\newcommand{\secondtxt}[1]{\colorbox{orange!15}{#1}}
\newcommand{\thirdtxt}[1]{\colorbox{yellow!15}{#1}}
\newcommand{\first}[1]{\cellcolor{red!15}#1}
\newcommand{\second}[1]{\cellcolor{orange!15}#1}
\newcommand{\third}[1]{\cellcolor{yellow!15}#1}

\definecolor{cvprblue}{rgb}{0.21,0.49,0.74}
\usepackage[pagebackref,breaklinks,colorlinks,citecolor=cvprblue]{hyperref}
\usepackage{xcolor}  
\usepackage{colortbl}       
\usepackage{amsmath}
\usepackage{pifont}

\def\paperID{20} 
\def\confName{3DV\xspace}
\def\confYear{2026\xspace}

\title{RGS-DR: Deferred Reflections and Residual Shading in 2D Gaussian Splatting}

\author{
{Georgios Kouros \qquad Minye Wu \qquad Tinne Tuytelaars} \\
{Department of Electrical Engineering (ESAT), KU Leuven, Belgium} \\
{\tt\small \{georgios.kouros,minye.wu,tinne.tuytelaars\}@esat.kuleuven.be}
}

\begin{document}
\maketitle
\begin{abstract}

In this work, we address specular appearance in inverse rendering using 2D Gaussian splatting with deferred shading and argue for a refinement stage to improve specular detail, thereby bridging the gap with reconstruction-only methods.
Our pipeline estimates editable material properties and environment illumination while employing a directional residual pass that captures leftover view-dependent effects for further refining novel view synthesis.
In contrast to per-Gaussian shading with shortest-axis normals and normal residuals, which tends to result in more noisy geometry and specular appearance, a pixel-deferred surfel formulation with specular residuals yields sharper highlights, cleaner materials, and improved editability.
We evaluate our approach on rendering and reconstruction quality on three popular datasets featuring glossy objects, and also demonstrate high-quality relighting and material editing. 
The source code is available at \url{https://github.com/gkouros/RGS-DR}.

\end{abstract}
\section{Introduction}

Reconstructing and rendering glossy and reflective objects remain significant challenges in computer vision and graphics. These objects exhibit strong view-dependent effects, such as specular highlights and reflections, which undermine multi-view consistency and increase the difficulty of recovering accurate 3D geometry and appearance from multi-view images. In real-world scenarios, such objects are prevalent, including automobiles, metallic artifacts, and ceramic tableware. Achieving high-quality reconstruction and rendering of such surfaces is crucial for enhancing realism and user immersion in applications such as virtual reality (VR), augmented reality (AR), and gaming.

\begin{figure}[t]
    \includegraphics[width=\linewidth]{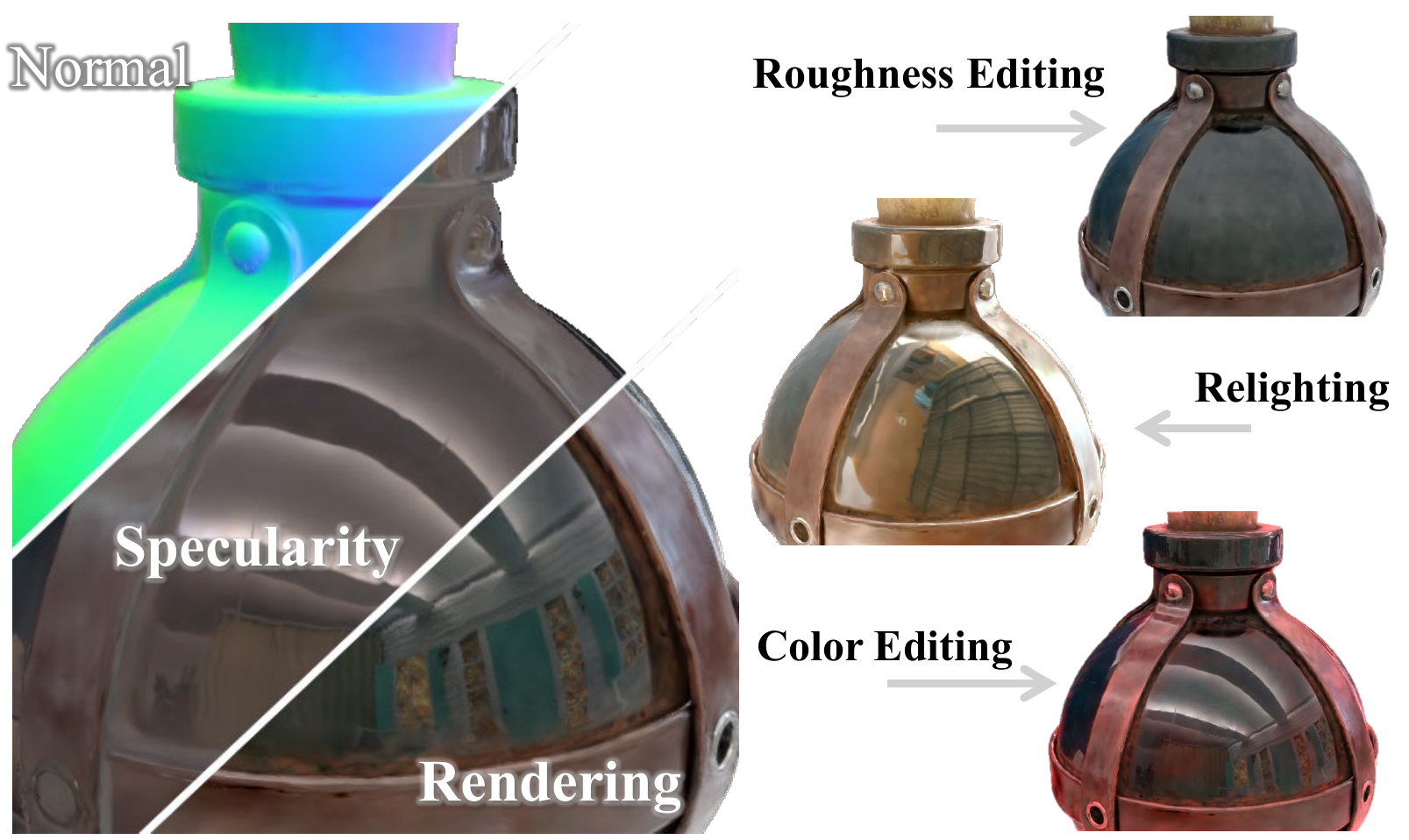}
    \vspace{-0.6cm}
    \caption{RGS-DR delivers high-quality inverse rendering by disentangling scene geometry, material properties, and illumination, enabling photorealistic rendering, relighting, and scene editing.}
    \label{fig:teaser}
\end{figure}

Recent advances in novel view synthesis (NVS) methods have demonstrated high-fidelity scene reconstruction and rendering quality when applied to objects with predominantly Lambertian reflectance. However, state-of-the-art techniques such as Neural Radiance Fields (NeRF)~\cite{mildenhall2021nerf} and 3D Gaussian Splatting~\cite{kerbl20233dgs} struggle to achieve satisfactory results on glossy and reflective surfaces. The primary limitation arises from their lack of explicit modeling of light transport. These methods rely heavily on multi-view consistency, which fails to capture sufficient cues for reconstructing complex specular effects. Consequently, artifacts such as blurring and floating structures (floaters) frequently occur.

To address these issues, inverse neural rendering aims to decompose scene properties and approximate light transport. Techniques in this area~\cite{verbin2022refnerf, mai2023neural, liang2023envidr, jiang2024gshader,ye2024gsdr, zhang2024refgs, yao2025refGS} have led to significant improvements in rendering shiny objects. A core component of these methods is the estimation of reflection directions based on surface normals, which is essential for modeling specular reflections under environmental lighting. The quality of reconstruction and rendering is highly dependent on the accuracy of these surface normals. To obtain them, Ref-NeRF~\cite{verbin2022refnerf} uses multilayer perceptrons to learn coordinate-based normals, while GaussianShader~\cite{jiang2024gshader} employs learnable normal residuals conditioned on the viewing direction. However, both methods detach the normal estimation from the underlying geometry, which may result in noisy normals. Other methods, such as NMF~\cite{mai2023neural} and ENVIDR~\cite{liang2023envidr}, achieve better performance by using normals derived from the gradient of the geometry, while 3DGS-DR~\cite{ye2024gsdr} relies on the shortest axis of each Gaussian ellipsoid to estimate normals that may or may not align with the actual surface of the object. In contrast, Ref-GS \cite{zhang2024refgs} takes advantage of standard surfel modeling~\cite{pfister2000surfels}, adopting 2D-oriented disks as surface elements, providing an explicit normal representation that enhances the modeling of shiny objects. However, Ref-GS \cite{zhang2024refgs} encodes some of the object's material properties implicitly together with the ambient lighting, thus preventing relighting or other scene editing tasks.

To achieve relighting with inverse rendering, it is essential to explicitly model the environment lighting within the rendering equation. This explicit modeling enables the seamless replacement of the original lighting with the desired target lighting for effective relighting. Existing methods approximate the rendering equation using different formulations to enable this explicit modeling. A key challenge in this process is the accurate modeling of the integral of incident light from different directions. Spherical Harmonics~\cite{mai2023neural} and Spherical Gaussians~\cite{verbin2022refnerf, zhang2021physg} are commonly used to represent the distribution of environment lighting, as they provide closed-form solutions for approximating light integration. Other methods \cite{jiang2024gshader,ye2024gsdr,yao2025refGS} utilize discrete grid textures to capture high-frequency signals in environment lighting. However, 3DGS-DR~\cite{ye2024gsdr} fails to properly formulate the integral using only a single-level environment map and does not learn explicit material properties, which limit its expressive capability and editability. Meanwhile, GaussianShader~\cite{jiang2024gshader}, which samples from multi-level mipmaps to approximate the integral, struggles with geometric uncertainty, resulting in degraded results. Ref-Gaussian~\cite{yao2025refGS} reduces geometric uncertainty by utilizing the shortest axes of 2D Gaussians, but struggles to decompose diffuse from specular appearance and has to fit the diffuse color with spherical harmonics, disentangled from base color, leading to incorrect estimation of the latter.

In this paper we present \textit{RGS-DR}, an inverse-rendering method for glossy objects explicitly scoped to direct image-based lighting (IBL). As in 2DGS~\cite{huang20242dgs} and related work~\cite{zhang2024refgs,yao2025refGS,tong2025gs2dgs}, we use surfels to obtain reliable geometry and normals. Scene geometry and physically based rendering (PBR) parameters are optimized on each 2D Gaussian and rasterized into an image-space material buffer via a pixel-deferred pipeline. Per-pixel shading with split-sum IBL improves the discrete approximation of the specular integral and avoids per-Gaussian alpha-blend artifacts typical of forward shading~\cite{jiang2024gshader}.
To better capture view-dependent detail, we add a lightweight directional residual in image space (Ref-GS encoding~\cite{zhang2024refgs}) that fits leftover specular effects such as glints and micro-geometry. Relighting and scene edits are then performed by swapping the estimated environment map and adjusting the recovered material maps (e.g., roughness, diffuse albedo).
Empirically, RGS-DR achieves competitive reconstruction under training illumination while providing consistent environment and material estimates, and it supports high-quality relighting and editing on datasets with highly reflective objects. Our contributions can be summarized as follows:
\begin{itemize}
    \item A 2DGS-based, inverse-rendering pipeline with deferred reflections that recovers editable materials and environment illumination.
    \item A refinement stage with a directional residual module that sharpens specular appearance for novel view synthesis and reduces the gap to reconstruction-only baselines.
    \item A detailed evaluation of rendering, reconstruction, relighting, and material editing, compared to (direct IBL) inverse rendering and reconstruction-only baselines.
\end{itemize}
\section{Related Work}

\vspace{2mm}
\indent
\textbf{Novel View Synthesis.}  
Neural Radiance Fields~(NeRF) \cite{mildenhall2020nerf} offer an end-to-end framework for synthesizing photorealistic images from sets of 2D images using implicit neural representations. While NeRF offers high photorealism and compact representation, it is inefficient in training and rendering. To address this, subsequent approaches have incorporated hybrid implicit-explicit representations, such as voxel grids~\cite{SunSC22,sun2022improved,liu2020neural,yu_and_fridovichkeil2021plenoxels} and hash encodings~\cite{mueller2022instant}, which significantly enhance computational efficiency without compromising reconstruction quality. 
However, these methods face challenges when dealing with shiny objects due to geometric ambiguities arising from multi-view inconsistencies and their stochastic geometry modeling. 

3D Gaussian Splatting (3DGS)~\cite{kerbl20233dgs} has recently emerged as a promising alternative to NeRF for 3D scene representation, delivering high-quality novel view synthesis and rapid rendering speeds. Advancements in 3DGS, such as the use of structured Gaussians~\cite{lu2024scaffold,chen2024pgsr, ren2024octree}, texture mapping~\cite{chao2024textured}, gradient cues~\cite{zhang2024pixel,rota2024revising}, and anti-aliasing~\cite{yu2024mip}, have further enhanced rendering quality. However, these methods often fall short in accurately reconstructing and rendering glossy or reflective surfaces due to their inability to precisely model surface-light interactions. To address these limitations, 2D Gaussian Splatting (2DGS)~\cite{huang20242dgs} projects Gaussian disks onto object surfaces and applies local smoothing. This approach ensures view-consistent geometry. Our method and other recent methods \cite{zhang2024refgs,yao2025refGS,tong2025gs2dgs} leverage its explicit surface normals combined with a tailored shading function to accurately model light transport in scenes featuring shiny objects.

\vspace{2mm}
\textbf{Inverse Neural Rendering. } 
Inverse rendering is a challenging problem that seeks to reconstruct scene properties, including geometry, materials, and lighting, from images. Recent advances in inverse rendering have integrated differentiable volume rendering with implicit neural representations, drawing inspiration from NeRF \cite{mildenhall2020nerf}, to learn scene representations encompassing both geometry and materials. Early approaches in this area often made strong assumptions or required additional priors, such as a known illumination \cite{srinivsan2021nerv, bi2020deepreflectancevolumes}, geometry \cite{zhang2021nerfactor}, or varying illumination conditions \cite{bi2020deepreflectancevolumes, gao2020deferred}. These limitations have been addressed in subsequent works \cite{zhang2021physg, boss2021nerd, boss2021neuralpil, liang2023envidr, fan2023factoredneus, Jin2023TensoIR, verbin2022refnerf, liu2023nero}, which jointly estimate geometry, materials (e.g., BRDF), and lighting. 
%

Gaussian-based neural rendering methods~\cite{jiang2024gshader, tang20243igs, wu2024deferredgs, zhu2024gs-sdf} have emerged as faster, more specular-aware alternatives to field-based NeRFs by explicitly computing reflection directions and using BRDF-aware shading. Several recent works~\cite{zhang2024refgs, ye2024gsdr, yao2025refGS, tong2025gs2dgs} adopt a deferred formulation~\cite{deering1988triangle}, rasterizing material buffers and shading in image space. We follow this pixel-deferred paradigm and explicitly estimate PBR material maps together with an HDR environment, performing split-sum IBL per pixel. This avoids per-primitive alpha-blended shading, which tends to smear speculars and couple errors to depth ordering (see Fig.~\ref{fig:teaser}). 
Closest to our setting, GaussianShader~\cite{jiang2024gshader} attaches a per-Gaussian residual color modeled with low-order spherical harmonics to absorb unmodeled view dependence. In contrast, we employ a deferred residual pass with the more expressive spherical directional encoding of Ref-GS~\cite{zhang2024refgs}. Residuals are predicted in image space (not stored per primitive), so we keep only a small per-Gaussian feature vector (much smaller than SH) for conditioning, yielding fewer parameters and lower memory at comparable cost. The residual is disabled at relight time to prevent training-illumination bias.
Several methods \cite{liu2023nero, yao2025refGS, younes2025texturesplat, chen2025gigs} extend split-sum IBL with additional components for indirect lighting, aiming to capture complex global illumination effects. However, these enhancements fall outside the scope of direct IBL, which is our focus. Other works \cite{li2025recap, tong2025gs2dgs, xie2024envgs} adopt fundamentally different assumptions, such as multi-light supervision or external geometric priors, making them incompatible with our setting. These directions target complementary problems and are not directly comparable to ours.
\section{Preliminary}

2D Gaussian Splatting~\cite{huang20242dgs} projects 3D Gaussians~\cite{kerbl20233dgs} onto 2D oriented planar disks, ensuring view-consistent geometry and explicit normals, defined as the direction of the steepest density change.
Each planar disk serves as a 2D Gaussian primitive, characterized by a center point $\mathbf{x}$, two principal tangential vectors $\mathbf{t}_u$ and $\mathbf{t}_v$, and two variance-controlling scaling factors $s_u$ and $s_v$ that need to be optimized. We use the covariance matrix $\Sigma$ to represent the rotation and scaling of each Gaussian.
Instead of evaluating Gaussian values at the intersection of a pixel ray and a 3D Gaussian, 2DGS computes them directly on 2D disks, leveraging explicit ray-splat intersections for perspective-correct splatting. The Gaussian value is formulated as:
$
\mathcal{G}(\mathbf{u}(\mathbf{r})) = \exp\!(-(u^2 + v^2)/{2})
$, 
where $\mathbf{u}(\mathbf{r})=(u,v)$ is the intersection point between
ray $\mathbf{r}$ and the disk in UV space.

For rendering, Gaussians along with their properties, such as color $\mathbf{c}_i$, are sorted by their centers and blended into pixels front to back, which is formulated by alpha blending: 
\begin{equation} \label{eq:alpha_blending}
c(\mathbf{r}) 
= \sum_{i=1}^{n} 
   \mathbf{c}_{i}\,\alpha_{i}\,\mathcal{G}_{i}\bigl(\mathbf{u}(\mathbf{r})\bigr)
   \prod_{j=1}^{i-1}\!\Bigl(1 - \alpha_{j}\,\mathcal{G}_{j}\bigl(\mathbf{u}(\mathbf{r})\Bigr),
\end{equation}
where $i$ and $\alpha_i$ denote the index of the Gaussian and its alpha value; $\mathbf{c}_i$ is the view-dependent appearance color for the $i$-th Gaussian. In our method, we need to render different object material properties (e.g. diffuse color, roughness, and normal) and feature vectors into the target view. This is achieved using alpha blending, where the principle is to replace the color 
$\mathbf{c}_i$ with the corresponding material properties or features, generating multi-channel pixel values.

Planar disks have well-defined surface normals that are perpendicular to their surfaces. Using the tangential vectors, we can compute the normal vectors in closed form: 
\begin{equation}\label{eq:normal}
\mathbf{n}=\mathbf{t}_u \times \mathbf{t}_v, 
\end{equation} 
where $\mathbf{n}$ is the  normal vector, and operator $\times$ stands for cross product. In our method, we use this explicit normal formulation in the light transport modeling, enabling accurate reflection calculations. This perspective-correct rendering improves rendering performance based on deferred shading.


\section{Methodology}
\begin{figure*}[t]
    \centering
    \includegraphics[width=\linewidth]{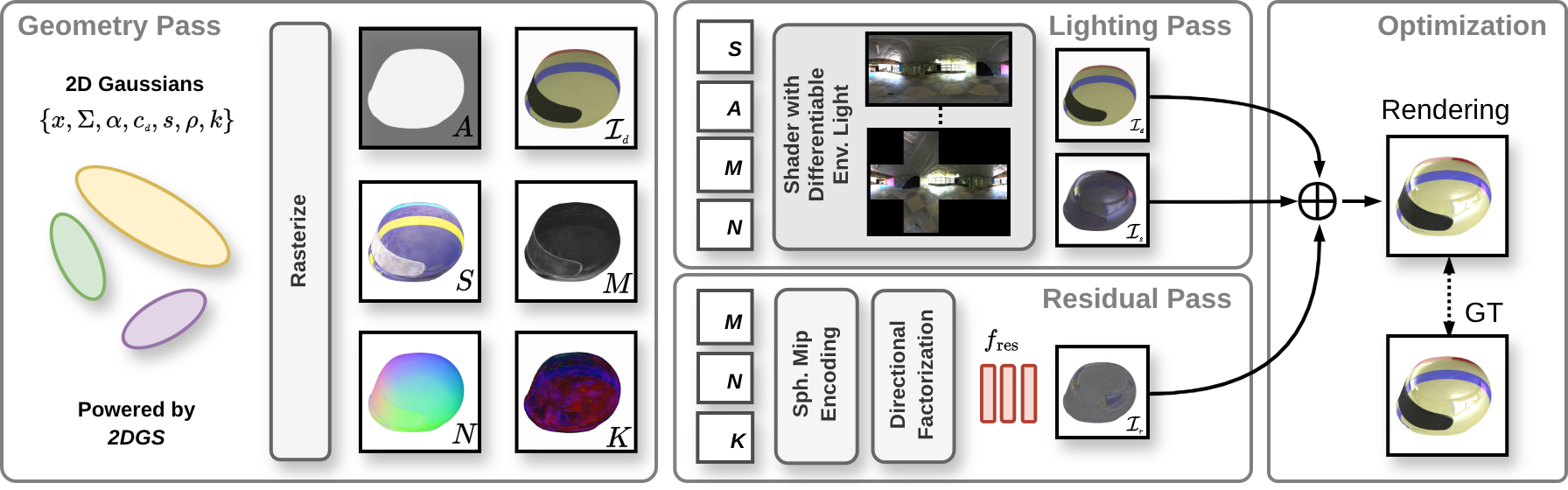}
    \vspace{-0.6cm}
    \caption{Our rendering pipeline consists of three passes. The geometry pass produces screen-space diffuse color $\mathcal{I}_d$, specular tint $S$, roughness $M$, normals $N$, and low-dimensional features $K$, which feed into the subsequent passes. The lighting pass employs a cube mipmap to model environmental light for shading as described in \cite{jiang2024gshader}. Meanwhile, the residual pass uses a spherical-mip-based directional encoding (inspired by \cite{zhang2024refgs}) along with a shallow MLP $f_\text{res}$ to predict view-dependent effects not captured by the lighting pass.}
    \label{fig:architecture}
\end{figure*}

%

RGS-DR exploits surfel primitives that have accurate surface properties to compose the scene. In addition to the geometry-related variables mentioned above, each primitive has learnable properties, namely diffuse color $\mathbf{c}_d$, roughness $\rho$, and specular tint $\mathbf{s}$.  We also assign a feature vector $\mathbf{f}$ to each primitive to capture inter-reflection. 

\vspace{2mm}
\textbf{Overview.} Instead of first calculating the view-dependent color of primitives and then blending them, we employ a deferred rendering scheme. Our approach rasterizes the geometry and blends the properties into a Graphics Buffer (G-Buffer), which stores buffer images for each property~(\S~\ref{sec:geometry_pass}). The G-Buffer stores essential information for modeling light transport in reflections, including geometry, material, and lighting-related properties. It enables a representation of global illumination effects. Leveraging this data, shading functions are applied to approximate the rendering equation based on the material attributes in the G-Buffer. This process enables the computation of pixel-wise colors for the specular component while accounting for indirect lighting contributions from the surrounding environment~(\S~\ref{sec:shading}). In addition to environmental illumination, light undergoes multiple reflections between object surfaces, a phenomenon known as inter-reflection, which is not explicitly modeled in standard shading functions. To address this limitation, we propose a residual rendering pass that integrates features from the G-Buffer with a spherical feature mip-map~\cite{zhang2024refgs} to effectively encode inter-reflection effects~(\S~\ref{sec:residual}).
We integrate these components into our proposed differentiable pipeline to generate the final rendered images. The complete pipeline is illustrated in Fig.~\ref{fig:architecture}. 

\vspace{2mm}
\textbf{Deferred Rendering.} RGS-DR adopts the rendering equation approximation of GaussianShader~\cite{jiang2024gshader} defined as:
\begin{equation}\label{rendering_eq}
\mathbf{c}(\boldsymbol{\omega}_o) = \mathbf{c}_d + \mathbf{s} \odot \mathbf{L}_s(\boldsymbol{\omega}_o, \mathbf{n}) + \mathbf{c}_r(\boldsymbol{\omega}_o),
\end{equation}
where $\boldsymbol{\omega}_o \in \mathbb{R}^3$ is the viewing direction, $\mathbf{c}_d \in \mathbb{R}^3$ is the diffuse color,
$\mathbf{s} \in \mathbb{R}^3$ is the specular tint,
$\mathbf{L}_s \in \mathbb{R}^3$ is the specular light from the shading functions~(\S~\ref{sec:shading}), 
$\rho \in \mathbb{R}^1$ is the roughness, and
$\mathbf{c}_r \in \mathbb{R}^3$ is a residual color term (\S~\ref{sec:residual}) that accounts for inter-reflection effects.  Note that deferred rendering computes on the image pixel level; all the input variables are fetched from the G-buffer at the corresponding pixel coordinates. 
The final rendered image $\mathcal{I}$ is calculated as
\begin{equation}\label{total_rendering_eq}
\mathcal{I} = \mathcal{I}_d + \mathcal{I}_s + \mathcal{I}_r,
\end{equation}
where $\mathcal{I}_d$ and $\mathcal{I}_s$ are the diffuse and specular components of the scene, while $\mathcal{I}_r$ is the residual image term. The components of Eq.~\ref{rendering_eq}-\ref{total_rendering_eq} will be introduced in the following subsections.


\subsection{G-Buffer Rendering} \label{sec:geometry_pass}
We adapt the alpha blending formulation in Eq.~\ref{eq:alpha_blending} to generate buffer images by replacing $\mathbf{c}_i$ with the corresponding property vectors. 
Following the notation of Ref-GS\cite{zhang2024refgs}, RGS-DR fills the G-buffer with the diffuse map $\mathcal{I}_d$, roughness map $\mathbf{M}$, specular tint map $\mathbf{S}$, feature map $\mathbf{K}$, and normal map $\mathbf{N}$, as well as the accumulated opacity $\mathbf{A}$. They have the same resolution as the render target but differ in the number of channels. Most of them are alpha-blended from corresponding properties stored in each Gaussian primitive except for the normal map $\mathbf{N}$, which is derived from the primitive geometry using Eq.~\ref{eq:normal}.  


\subsection{Deferred Shading Functions} \label{sec:shading}


Shading functions are an important part of the approximation of the rendering equation~\cite{kajiya1986rendering} to model light-surface interactions. They have been used in \cite{jiang2024gshader}, demonstrating their effectiveness in material decomposition and representation. In this work, we extend shading functions to the deferred rendering framework and apply them to the G-buffer.

We use a shading function to compute the specular component $\mathbf{L}_s$, which is formulated as: 
\begin{equation}
\mathbf{L}_s(\boldsymbol{\omega}_o, \mathbf{n})
= \int_{\Omega}
L(\boldsymbol{\omega}_i)\,f_{r}(\boldsymbol{\omega}_i,\boldsymbol{\omega}_o)\,\bigl(\boldsymbol{\omega}_i \cdot \mathbf{n}\bigr)\,d\boldsymbol{\omega}_i,
\end{equation}
%
where $\boldsymbol{\omega}_i$ is the direction of the input radiance, $\Omega$ denotes the directions of the entire upper hemisphere above the current primitive disk, and $f_r$ is the bidirectional reflectance distribution function.
    
We use the split-sum approximation of the integral to avoid computationally expensive explicit integral evaluations. Specifically, we achieve this using a pre-filtered cube mip-map $\mathbf{E}$ where each mip-map level stores progressively blurrier versions of the original environment map, simulating different levels of surface roughness for reflections.

In contrast to GaussianShader~\cite{jiang2024gshader} which applies shading on each individual 3D Gaussian before splatting, we defer shading and apply it at the pixel level. This approach provides two benefits. First, we compute lighting only on the visible parts of the scene, and second, we leverage the smooth normals that are produced from 2D Gaussian primitive, which are far more accurate than the shortest normal axes of 3D Gaussians. At the same time, we do not have to learn additional normal residuals as does GaussianShader to account for inconsistencies between surface normals and Gaussian normals.  
In the deferred rendering scheme, pixel shading is performed without alpha blending, resulting in high-frequency reflections. 


\subsection{Residual Rendering Pass}\label{sec:residual}

We incorporate a residual color term optimized in what we call the residual pass, as shown in Fig.\ref{fig:architecture}. 
Instead of using high-dimensional spherical harmonic coefficients per Gaussian, we employ low-dimensional features $\mathbf{k}$ and optimize a directional encoding based on the methodology of
Ref-GS \cite{zhang2024refgs}. This involves a spherical mipmap-based directional encoding that enables roughness-aware photorealistic rendering on highly reflective scenes. However, the environment illumination in Ref-GS is baked in the mipmap, and thus learned scenes are not relightable. Instead, we leverage the high expressivity of such a directional encoding to learn scene-specific information and augment the reconstruction quality of our model. 
The resulting residual color term complements our explicit shading branch and accounts for missing complex view-dependent effects, such as inter-reflections, thus refining our method's reconstruction quality. However, relighting quality is not affected as the residual term is illumination-specific and is not included when performing relighting by rendering with new environment maps. 

For a given screen-space point $\mathbf{u}(u,v)$, the spherical mip encoding can be calculated based on the viewing direction $\boldsymbol{\omega}_o$, normal $\mathbf{n}$, roughness $\rho$, and feature $\mathbf{k}$. 
The viewing direction and normal are used to estimate the reflected direction $\boldsymbol{\omega}_r$ which is then converted into polar coordinate $(\theta_r,\phi_r)$.
The spherical mip encoding is then computed by trilinear interpolation on the spherical mipmap $\mathcal{M}$ along the roughness dimension and given by $\mathbf{h} = \mathrm{Sph\text{-}Mip}\bigl(\boldsymbol{\omega}_{r}, \rho, \mathcal{M}\bigr)$. 
This feature is concatenated with the outer product $\mathbf{h} \otimes \mathbf{k}$ to form the directional encoding, which is used as input to a shallow MLP $f_\text{res}$ to produce the residual color term $\mathbf{c}_r = f_\text{res}(\mathbf{h}, \mathbf{k} \otimes \mathbf{h})$ and the corresponding residual image 
\begin{equation}
\mathcal{I}_r = \textit{f}_\text{res}(\mathbf{H}, \mathbf{K} \otimes \mathbf{H}).
\end{equation}


\subsection{Training Strategy}

To train our model, we follow the training strategy of 3DGS \cite{kerbl20233dgs}, which supervises the model with a photometric RGB reconstruction loss $\mathcal{L}_c$ that combines an $\mathcal{L}_1$ loss with a D-SSIM term into

\begin{equation}
\mathcal{L}_c = (1 - \lambda)\,\mathcal{L}_1 + \lambda\,\mathcal{L}_{\mathrm{D\text{-}SSIM}},
\end{equation}
where $\lambda$ is the balancing term between the two losses and is usually chosen as $\lambda=0.2$.

However, the problem of estimating 3D geometry from a collection of posed 2D images using only a photometric loss is ill-posed. As a result, we additionally adopt the regularization methodology of 2DGS \cite{huang20242dgs}. This regularization involves the normal consistency regularization from Ref-NeRF \cite{verbin2022refnerf}, given in Eq. \ref{eq:normalloss} and the depth distortion regularization from MipNeRF360 \cite{barron2022mip360}, given in eq. \ref{eq:distloss}. The former tries to enforce consistency between the predicted normal map and the normals computed by depth gradients and is formulated as

\begin{equation}\label{eq:normalloss}
\mathcal{L}_n = \sum_{i} w_i \bigl(1 - \mathbf{n}_i^{T}\mathbf{N}\bigr),
\end{equation}
where $i$, $w_i$, $\mathbf{n}_i$, and $N$ correspond to a ray-splat intersection, its blending weight, its normal, and the local depth normal calculated using finite differences from nearby depth points. 
The latter regularizes the weight distribution of the splats to concentrate them around the surface by minimizing the surface-splat intersection. This is accomplished through  
\begin{equation}\label{eq:distloss}
\mathcal{L}_d = \sum_{i,j} w_i \,w_j \,\bigl|z_i - z_j\bigr|,
\end{equation}
where $z_i$ is the depth of the \textit{i}-th ray-splat intersection.

Furthermore, we use an alpha loss to prevent floaters caused by redundant Gaussians in synthetic scenes: 
\begin{equation}\label{eq:sillhouette}
\mathcal{L}_{\mathrm{\alpha}} = {\bigl\lVert A-A_{gt} \bigr\rVert}_1,
\end{equation}
where $A$ is the accumulated alpha map in the G-buffer and $A_{gt}$ is the corresponding ground-truth alpha channel of the provided RGBA image. 

The total loss is calculated as a weighted sum of the individual
loss terms as
\begin{equation}
\mathcal{L} = \mathcal{L}_{c} + \lambda_d \mathcal{L}_{d} + \lambda_n \mathcal{L}_{n} + \lambda_\alpha \mathcal{L}_\alpha,
\end{equation}
where $\lambda_d$, $\lambda_n$, and $\lambda_{\alpha}$ are the corresponding loss weights.

\section{Experimental Results}

\subsection{Datasets}
We compare our method against state-of-the-art baselines on two synthetic datasets and one real dataset with complex, view-dependent effects resulting from highly reflective surfaces. Specifically, we evaluate on the Shiny Synthetic~\cite{verbin2022refnerf}, Glossy Synthetic~\cite{liu2023nero}, and Shiny Real \cite{verbin2022refnerf}.
Results are provided for all scenes in the Shiny Synthetic and Shiny Real datasets. For the Glossy Synthetic dataset, we select the same subset of scenes as in \cite{ye2024gsdr, zhang2024refgs}. 

\subsection{Implementation Details}
All our models are trained on an NVIDIA Tesla P100 GPU, completing 35k iterations in two hours on average. For the lighting pass, we use an environment cubemap of size $6\times512\times512$ with 3 RGB channels, and for the residual pass an $8\times512\times512$ mipmap with 16 feature channels, along with a shallow MLP (two hidden layers of 256 units) plus one 4D feature per Gaussian.
For the real scenes, we downsample the training images of the gardenspheres and toycar scenes by factors of 4 and the sedan scene by a factor of 8 to be consistent with \cite{ye2024gsdr,zhang2024refgs}. Furthermore, we employ a similar bounding volume as in \cite{ye2024gsdr,zhang2024refgs} to allow for reflective surfaces only in the foreground object. We achieve this by penalizing low roughness properties in Gaussians outside of the bounding volume.
We use loss weights of $\lambda_n=0.05$, $\lambda_d=100$, and $\lambda_\alpha=1.0$. Material properties are optimized with a learning rate of $2.5\times10^{-3}$, while opacity, scale, and rotation use $0.03$, $3\times10^{-3}$, and $10^{-2}$, respectively. The environment cube mipmap is updated with an exponentially decaying rate from $10^{-2}$ to $10^{-3}$. Gaussian positions are trained as in 2DGS~\cite{huang20242dgs}, with a decay from $1.6\times10^{-4}$ to $1.6\times10^{-6}$.
We train 30k iters without the residual, then enable it and freeze geometry/materials/lighting for a 5k-iter fine-tune. Only the residual branch is updated, benefiting NVS but not affecting editing.

\subsection{Comparison}
\begin{figure}[t]
    \centering
    \includegraphics[width=\linewidth]{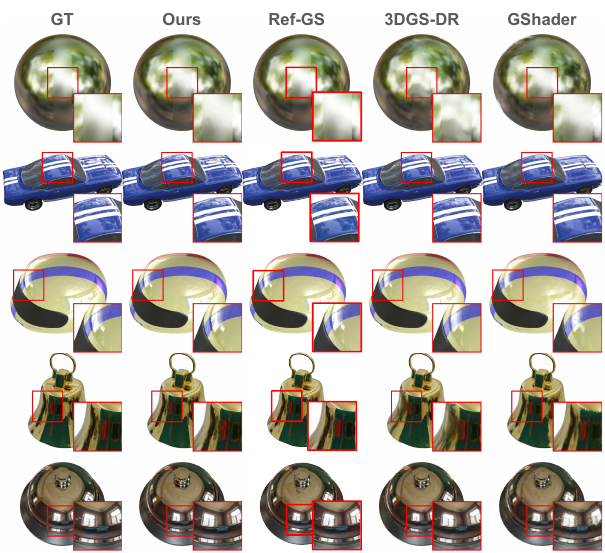}
    \vspace{-0.6cm}
    \caption{Comparison of the rendering quality of our method against Ref-GS~\cite{zhang2024refgs}, 3DGS-DR~\cite{ye2024gsdr} and GaussianShader~\cite{jiang2024gshader}.}
    \label{fig:comparisons}
\end{figure}

In Table~\ref{tab:comparisons} and Fig.~\ref{fig:comparisons}, we present the quantitative and qualitative evaluation of our proposed methodology against state-of-the-art novel view synthesis and inverse rendering methods. As shown in the table, we outperform the competing methods on several scenes of the examined datasets. Specifically, our method appears superior on the most glossy objects such as car, helmet, and toaster, as well as the entire Glossy Synthetic dataset. Our results in Fig.~\ref{fig:comparisons} highlight the ability of our method to reconstruct sharper reflections with less blurring and distortions than the two main competing inverse rendering methods.

\begin{table*}[ht]
\setlength{\tabcolsep}{1.6pt}
\centering
\caption{Quantitative results comparing our method with state-of-the-art approaches on test views from the Shiny Synthetic, Shiny Real, and Glossy Synthetic datasets. Metrics from \cite{zhang2024refgs,ye2024gsdr} are used, with the top three methods for each scene and metric marked as \firsttxt{first}, \secondtxt{second}, and \thirdtxt{third}. The final columns present the average scores for each dataset. The second column indicates whether a method supports relighting (\cmark) or is only reconstruction-capable (\xmark) .}
\label{tab:comparisons}
\renewcommand{\arraystretch}{1.0}  
\resizebox{\linewidth}{!}{%
\begin{tabular}{l|c|ccccccc|ccccccc|cccc}
\toprule
 & \textbf{Rel-} & \multicolumn{7}{c|}{\textbf{Shiny Synthetic \cite{verbin2022refnerf}}} 
 & \multicolumn{7}{c|}{\textbf{Glossy Synthetic \cite{liu2023nero}}}
 & \multicolumn{4}{c}{\textbf{Shiny Real \cite{verbin2022refnerf}}}\\
 & \textbf{ight} & \textbf{ball} & \textbf{car} & \textbf{coffee} & \textbf{helmet} & \textbf{teapot} & \textbf{toaster} & \textbf{\textit{avg}}
& \textbf{bell} & \textbf{cat} & \textbf{luyu} & \textbf{potion} & \textbf{tbell} & \textbf{teapot} & \textbf{\textit{avg}}
& \textbf{garden} & \textbf{sedan} & \textbf{toycar} & \textbf{\textit{avg}}\\
\midrule
\multicolumn{19}{c}{\textbf{PSNR} $\uparrow$}\\
\midrule
Ref-NeRF & \xmark
  & 33.16 & \third{30.44} & \third{33.99} & 29.94 & 45.12 & 26.12 & 33.13
  & 30.02 & 29.76 & 25.42 & 30.11 & 26.91 & 22.77 & 27.50
  & \third{22.01} & 25.21 & 23.65 & 23.62
  \\
3DGS & \xmark
  & 27.65 & 27.26 & 32.30 & 28.22 & 45.71 & 20.99 & 30.36
  & 25.11 & 31.36 & 26.97 & 30.16 & 23.88 & 21.51 & 26.50
  & 21.75 & 26.03 & 23.78 & 23.85
  \\
GShader & \cmark
  & 30.99 & 27.96 & 32.39 & 28.32 & 45.86 & 26.28 & 31.97 
  & 28.07 & 31.81 & 27.18 & 30.09 & 24.48 & 23.58 & 27.54
  & 21.74 & 24.89 & 23.76 & 23.46
  \\
ENVIDR & \cmark    
  & \first{41.02} & 27.81 & 30.57 & \third{32.71} & 42.62 & 26.03 & 33.46
  & 30.88 & 31.04 & 28.03 & 32.11 & 28.64 & \first{26.77} & 29.58
  & 21.47 & 24.61 & 22.92 & 23.00
  \\
3DGS-DR & \cmark
  & 33.66 & 30.39 & \first{34.65} & 31.69 & \first{47.12} & \third{27.02} & \third{34.09}
  & \third{31.65} & \first{33.86} & \third{28.71} & \second{32.79} & \third{28.94} & 25.36 & \third{30.22}
  & 21.82 & \second{26.32} & \third{23.83} & \third{23.99}
  \\
Ref-GS & \xmark
  & \third{36.10} & \second{30.94} & \second{34.38} & \second{33.40} & \second{46.69} & \second{27.28} & \second{34.80}
  & \second{31.70} & \second{33.15} & \second{29.46} & \third{32.64} & \second{30.08} & \third{26.47} & \second{30.58}
  & \second{22.48} & \first{26.63} & \second{24.20} & \second{24.44}
  \\
Ours & \cmark
  & \second{37.76} & \first{31.05} & 33.54 & \first{33.97} & \third{46.29} & \first{27.54} & \first{35.02}
  & \first{32.15} & \third{32.95} & \first{29.74} & \first{33.10} & \first{30.54} & \second{26.61} & \first{30.85}
  & \first{23.19} & \third{26.25} & \first{24.45} & \first{24.63}
  \\
\midrule
\multicolumn{19}{c}{\textbf{SSIM} $\uparrow$}\\
\midrule
Ref-NeRF & \xmark
  & 0.971 & 0.950 & \third{0.972} & 0.954 & \third{0.995} & 0.921 & 0.961 
  & 0.941 & 0.944 & 0.901 & 0.933 & 0.947 & 0.897 & 0.927
  & \second{0.584} & 0.720 & 0.633 & 0.646
  \\
3DGS & \xmark        
  & 0.937 & 0.931 & \third{0.972} & 0.951 & \second{0.996} & 0.894 & 0.947
  & 0.892 & 0.959 & 0.916 & 0.938 & 0.908 & 0.881 & 0.916
  & 0.571 & \third{0.771} & 0.637 & \third{0.660}
  \\
GShader & \cmark     
  & 0.966 & 0.932 & 0.971 & 0.951 & \second{0.996} & 0.929 & 0.958 
  & 0.919 & 0.961 & 0.914 & 0.938 & 0.898 & 0.901 & 0.922
  & 0.576 & 0.728 & 0.637 & 0.647
\\
ENVIDR & \cmark
  & \first{0.997} & 0.943 & 0.962 & \first{0.987} & \third{0.995} & \first{0.990} & \first{0.979} 
  & 0.954 & \third{0.965} & 0.931 & \second{0.960} & 0.947 & \first{0.957} & 0.952
  & 0.561 & 0.707 & 0.549 & 0.606
  \\
3DGS-DR & \cmark
  & 0.979 & \second{0.962} & \first{0.976} & 0.971 & \first{0.997} & 0.943 & 0.971
  & \third{0.962} & \first{0.976} & \third{0.936} & \third{0.957} & \third{0.952} & 0.936 & \third{0.953}
  & \third{0.581} & \second{0.773} & \third{0.639} & \second{0.664}
  \\
Ref-GS & \xmark 
  & \third{0.981} & \third{0.961} & \second{0.973} & \third{0.975} & \first{0.997} & \second{0.950} & \third{0.973} 
  & \first{0.965} & \second{0.973} & \second{0.946} & \third{0.957} & \second{0.956} & \third{0.944} & \second{0.957}
  & 0.507 & \first{0.783} & \second{0.682} & 0.657
  \\
Ours & \cmark
  & \second{0.988} & \first{0.967} & 0.971 & \second{0.977} & \first{0.997} & \third{0.946} & \second{0.974}
  & \second{0.964} & \second{0.973} & \first{0.952} & \first{0.964} & \first{0.969} & \second{0.951} & \first{0.962}
  & \first{0.638} & 0.769 & \first{0.687} & \first{0.698}
  \\
\midrule
\multicolumn{19}{c}{\textbf{LPIPS} $\downarrow$}\\
\midrule
Ref-NeRF & \xmark           
  & 0.166 & 0.050 & 0.082 & 0.086 & 0.012 & 0.083 & 0.080 
  & 0.102 & 0.104 & 0.098 & 0.084 & 0.114 & 0.098 & 0.100
  & 0.251 & 0.234 & \first{0.231} & 0.239
  \\
3DGS & \xmark             
  & 0.162 & 0.047 & \third{0.079} & 0.081 & 0.008 & 0.125 & 0.084
  & 0.104 & 0.062 & 0.064 & 0.093 & 0.125 & 0.102  & 0.092
  & \third{0.248} & \second{0.206} & \third{0.237} & \third{0.230}
  \\
GShader & \cmark    
  & 0.121 & 0.044 & \second{0.078} & 0.074 & \third{0.007} & \second{0.079} & 0.067
  & 0.098 & 0.056 & 0.064 & 0.088 & 0.122 & 0.091 & 0.087
  & 0.274 & 0.259 & 0.239 & 0.257
  \\
ENVIDR & \cmark    
  & \first{0.020} & 0.046 & 0.083 & \first{0.036} & 0.009 & \third{0.081} & \first{0.046}
  & \second{0.054} & 0.049 & 0.059 & \second{0.072} & \third{0.069} & \first{0.041} & \second{0.057}
  & 0.263 & 0.387 & 0.345 & 0.332
  \\
3DGS-DR & \cmark
  & \third{0.098} & \second{0.033} & \first{0.076} & \third{0.049} & \first{0.005} & \third{0.081} & \third{0.057} 
  & 0.064 & \second{0.040} & \third{0.053} & \third{0.075} & \second{0.067} & 0.067 & 0.061
  & \second{0.247} & \third{0.208} & \first{0.231} & \second{0.229}
  \\
Ref-GS &  \xmark    
  & \third{0.098} & \third{0.034} & 0.082 & \second{0.045} & \second{0.006} & \first{0.070} & \second{0.056}
  & \first{0.049} & \third{0.041} & \second{0.046} & 0.076 & 0.073 & \third{0.064} & \third{0.058}
  & \first{0.242} & \first{0.196} & \second{0.236} & \first{0.225}
  \\
Ours & \cmark
  & \second{0.080} & \first{0.032} & 0.084 & 0.050 & \third{0.007} & 0.097 & 0.059 
  & \third{0.060} & \first{0.038} & \first{0.042} & \first{0.057} & \first{0.049} & \second{0.050} & \first{0.049} 
  & 0.264 & 0.225 & 0.263 & 0.251
  \\
\bottomrule
\end{tabular}
} 
\end{table*}

\begin{figure*}[t]
    \centering
    \includegraphics[width=\linewidth]{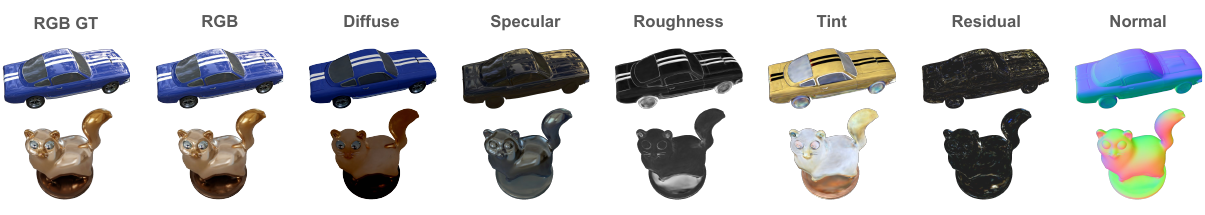}
    \caption{Outputs of our model including renderings, material properties, residuals and surface normals for scenes from the Shiny Synthetic and Glossy Synthetic datasets. The residuals are amplified for visualization purposes.}
    \label{fig:materials}
\end{figure*}

In Fig.~\ref{fig:materials}, we present the decomposition of our method's output, including the material properties, smooth geometry (visualized through surface normals), as well as the estimated specular components and residuals. As demonstrated, our method effectively disentangles the geometry, material properties, and illumination of a given scene, thus enabling photorealistic novel view synthesis, relighting, and scene editing. Unlike 3DGS-DR, our method learns explicit material properties and thus trained scenes can be easily ported to existing 3D rendering engines. 

\begin{figure*}[t]
    \centering
    \includegraphics[width=\linewidth,trim=0cm 0cm 0cm 0cm,clip]{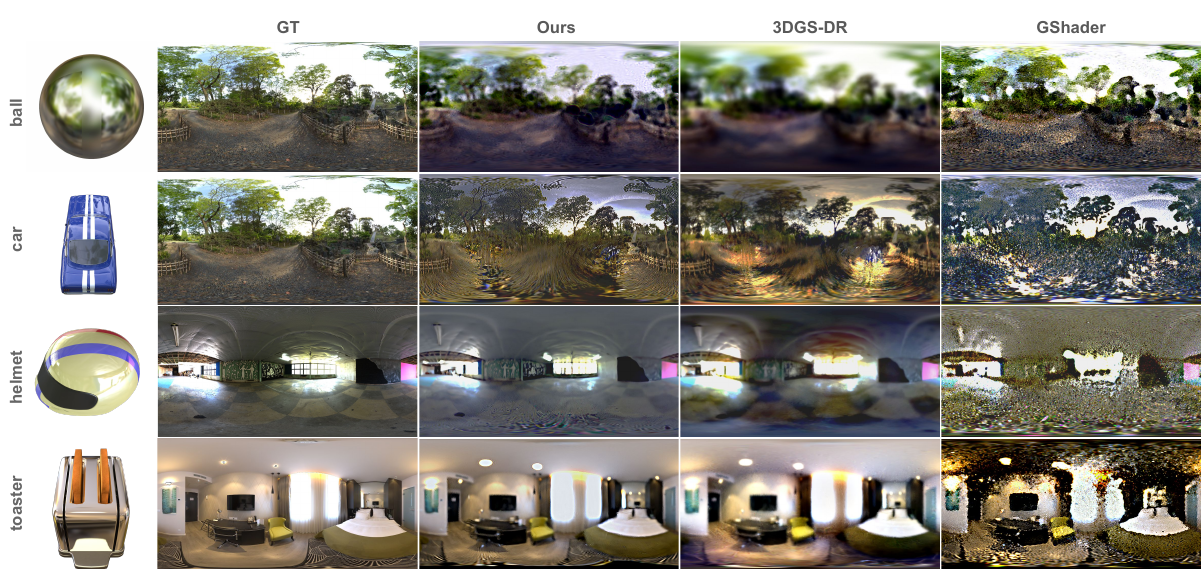}
    \vspace{-0.6cm}
    \caption{Comparison of estimated environment maps of synthetic scenes from the Shiny Synthetic dataset \cite{verbin2022refnerf}.}
    \label{fig:envmaps}
\end{figure*}

In Fig.~\ref{fig:envmaps}, we present our method's estimated environment illumination for scenes from the Shiny Synthetic dataset \cite{verbin2022refnerf}. Our method manages to effectively disentangle the illumination from the material properties and geometry of the scene. Compared to 3DGS-DR and GaussianShader, we obtain more refined environment maps with sharper details and fewer artifacts. A quantitative evaluation is provided in Table~\ref{tab:mae-emetrics}, containing the surface normal error and environment map quality for Shiny Synthetic.

\begin{table}[t]
\centering
\setlength{\tabcolsep}{3pt}
\caption{Quantitative comparison of surface normals via MAE$^\circ$ and environment map recovery via E-PSNR, E-SSIM, and E-LPIPS on the Shiny Synthetic dataset \cite{verbin2022refnerf}.}
\label{tab:mae-emetrics}
\begin{tabular}{l c c c c}
\hline
& MAE$^\circ$ $\downarrow$ & E-PSNR $\uparrow$ & E-SSIM $\uparrow$& E-LPIPS $\downarrow$ \\
\hline
GShader & 6.735 & 11.06 & 0.142 & 0.644 \\
3DGS-DR & 2.266 & 11.89 & 0.339 & 0.599 \\
Ref-GS & 2.210 & - & - & - \\
Ours & \textbf{1.917} & \textbf{13.28} & \textbf{0.403} & \textbf{0.563} \\
\hline
\end{tabular}
\end{table}

\begin{figure}[!t]
    \centering
    \includegraphics[width=\linewidth]{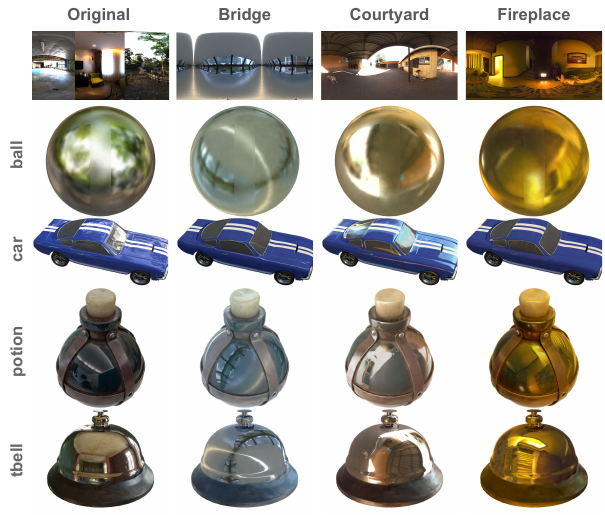}
    \vspace{-0.7cm}
    \caption{Relighted scenes with GT on the left and utilized environment map on the top.}
    \label{fig:relighting}
\end{figure}

\subsection{Ablation Study}
In Table~\ref{tab:ablation}, we conduct an ablation study to validate our design choices. First, we replace the cube mipmap of the environment light with a single cubemap to compare with 3DGS-DR which uses a single level. We observe a considerable drop in performance, demonstrating the necessity of a multi-resolution cubemap for accurate estimates of environment illumination. Second, we ablate the residual pass and once again notice a drop in performance, which highlights its benefit to photorealistic NVS. 
Finally, we replace the residual pass with spherical-harmonics-based residuals like in \cite{jiang2024gshader}, which improves performance but not as much as our proposed residual pass.


\begin{table}[t]
\centering
\caption{Ablation study on the Shiny Synthetic dataset \cite{verbin2022refnerf}. 
We ablate the environment map mipmapping, the residual pass, and also evaluate an alternative residual pass with spherical harmonics.
}
\label{tab:ablation}
\begin{tabular}{l c c c}
\hline
& PSNR\(\uparrow\) & SSIM \(\uparrow\) & LPIPS\(\downarrow\) \\
\hline
w/o mipmap          & 33.89 & 0.968 & 0.068 \\
w/o residual        & 34.28 & \textbf{0.974} & \textbf{0.058} \\
w/ SH-residual      & 34.68 & 0.973 & 0.059 \\
Ours                & \textbf{35.02} & \textbf{0.974} & 0.059 \\
\hline
\end{tabular}
\end{table}

\subsection{Relighting and Scene Editing}
In Fig.~\ref{fig:relighting}, we present the relighting capabilities of our model for two scenes from the Shiny Synthetic dataset \cite{verbin2022refnerf} and two scenes from the Glossy Synthetic dataset \cite{liu2023nero}. To relight the scenes, we replace the learned environment map and convert it to a cube mipmap before rendering. As demonstrated, our method can effectively relight learned scenes and achieve photorealistic results even under new illuminations. Additional relighting results can be found in the supplementary material. In Fig.~\ref{fig:editing}, on the other hand, we demonstrate the scene editing capabilities of our model on three glossy objects by modifying their roughness to make them rougher or smoother, and by changing their diffuse color.

\begin{figure}[t]
    \centering
    \includegraphics[width=\linewidth,trim=0cm 0cm 0cm 0cm,clip]{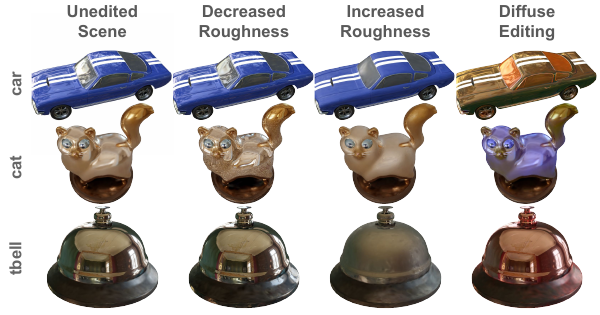}
    \vspace{-0.7cm}
    \caption{Demonstration of the material editing capabilities of our method.}
    \label{fig:editing}
\end{figure}
\section{Conclusion}

%

This work tackles inverse rendering of glossy objects with complex view-dependent effects using a 2D Gaussian, pixel-deferred formulation.
Our method estimates editable material properties and environment illumination via split-sum IBL, and adds a refinement stage with a lightweight directional residual that captures leftover specular detail, further enhancing novel view synthesis and bridging the gap with reconstruction-only methods.
Across challenging reflective scenes, we observe sharper highlights, plausible material estimates, and higher relighting fidelity than per-Gaussian shading baselines or uniform roughness baselines.
Our scope is limited to direct IBL since we do not model multi-bounce indirect lighting, transparency, or subsurface scattering. Scenes dominated by strong near-field interreflections are better addressed by global-illumination-enabled variants.
\section*{Acknowledgments}
\begin{sloppypar}
This work was supported by the Flanders AI Research Program and the KU Leuven Methusalem project "Lifelines".
The resources and services used in this work were provided by the VSC (Flemish Supercomputer Center), funded by the Research Foundation - Flanders (FWO) and the Flemish Government.
\end{sloppypar}
{
    \small
    \bibliographystyle{ieeenat_fullname}
    \bibliography{main}
}
\clearpage
\setcounter{page}{1}
\maketitlesupplementary

%

In this supplementary material, we provide additional qualitative results to showcase the contributions of our proposed methodology, including model outputs and scene decompositions, estimated environment maps, and relightings. Last but not least, we provide a video demonstration of our method with smooth animations of rendered scenes showcasing the capabilities of our method for reconstruction, relighting and scene editing. We compare our results with the corresponding results of 3DGS-DR \cite{ye2024gsdr} and Gaussian Shader \cite{jiang2024gshader} and demonstrate the superiority of our approach.

\section{Model Outputs and Scene Properties}
\begin{figure*}[t]
    \centering
    \includegraphics[width=0.98\linewidth]{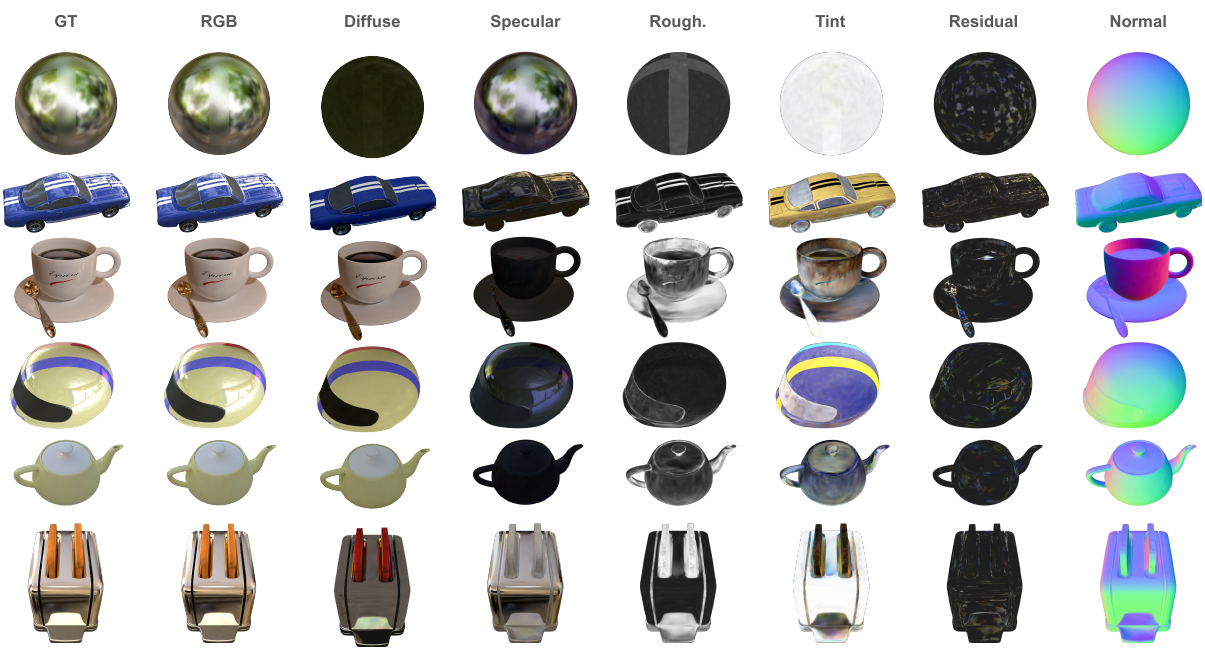}
    \caption{Outputs of our model including diffuse, specular and residual colors, material properties, as well as surface normals for scenes from the Shiny Synthetic \cite{verbin2022refnerf}. The residuals are amplified for visualization purposes.}
    \label{fig:shiny_materials}
\end{figure*}

\begin{figure*}[t]
    \centering
    \includegraphics[width=0.98\linewidth]{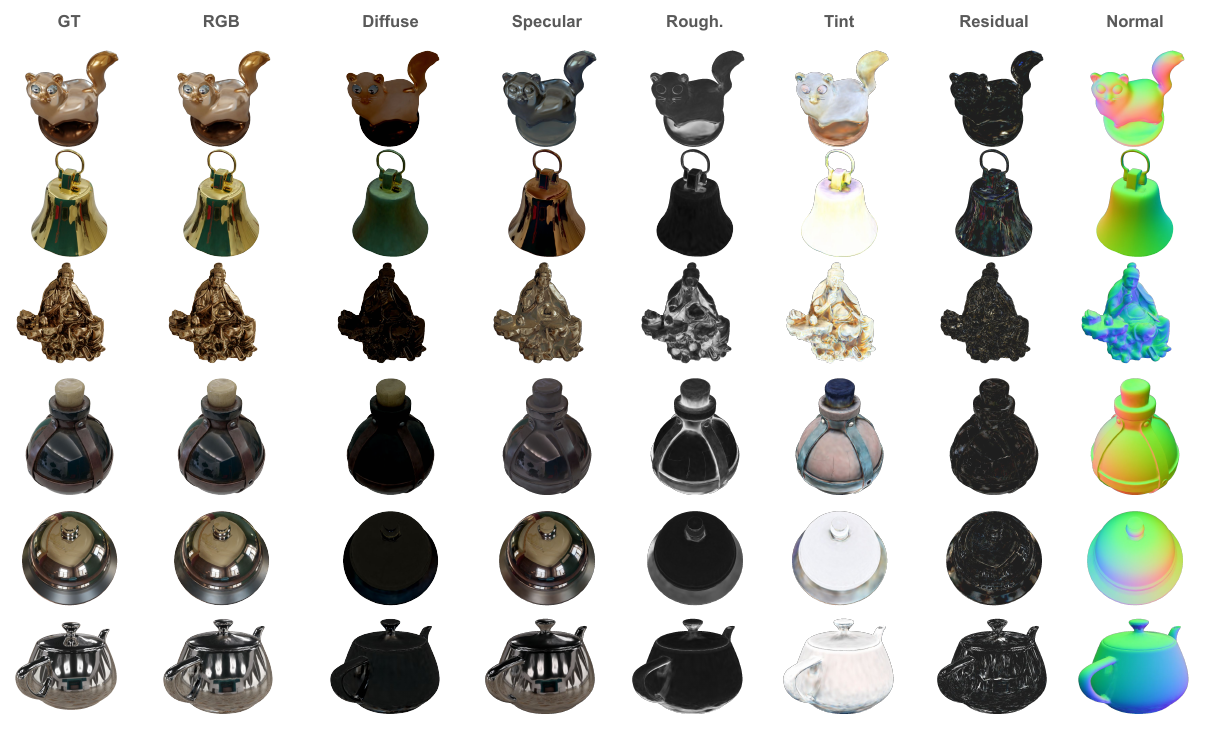}
    \caption{Outputs of our model including diffuse, specular and residual colors, material properties, as well as surface normals for scenes from the Glossy Synthetic \cite{liu2023nero}. The residuals are amplified for visualization purposes.}
    \label{fig:glossy_materials}
\end{figure*}

In Fig.~\ref{fig:shiny_materials} and Fig.~\ref{fig:glossy_materials}, we present 
the outputs and scene decompositions of our method on the Shiny Synthetic \cite{verbin2022refnerf} dataset and the Glossy Synthetic \cite{liu2023nero} dataset, respectively. From left to right, we present the ground-truth color, our method's reconstructed appearance, diffuse color, specular color, roughness, specular tint, residual color and surface normals. These results demonstrated the capability of our method to effectively disentangle the geometry, material properties and illumination during inverse rendering thus facilitating high-quality reconstruction, relighting, and scene editing of reflective surfaces.



\section{Recovered Environment Maps}
\begin{figure*}[t]
    \centering
    \includegraphics[width=0.82\linewidth,trim=0cm 0cm 0cm 0cm,clip]{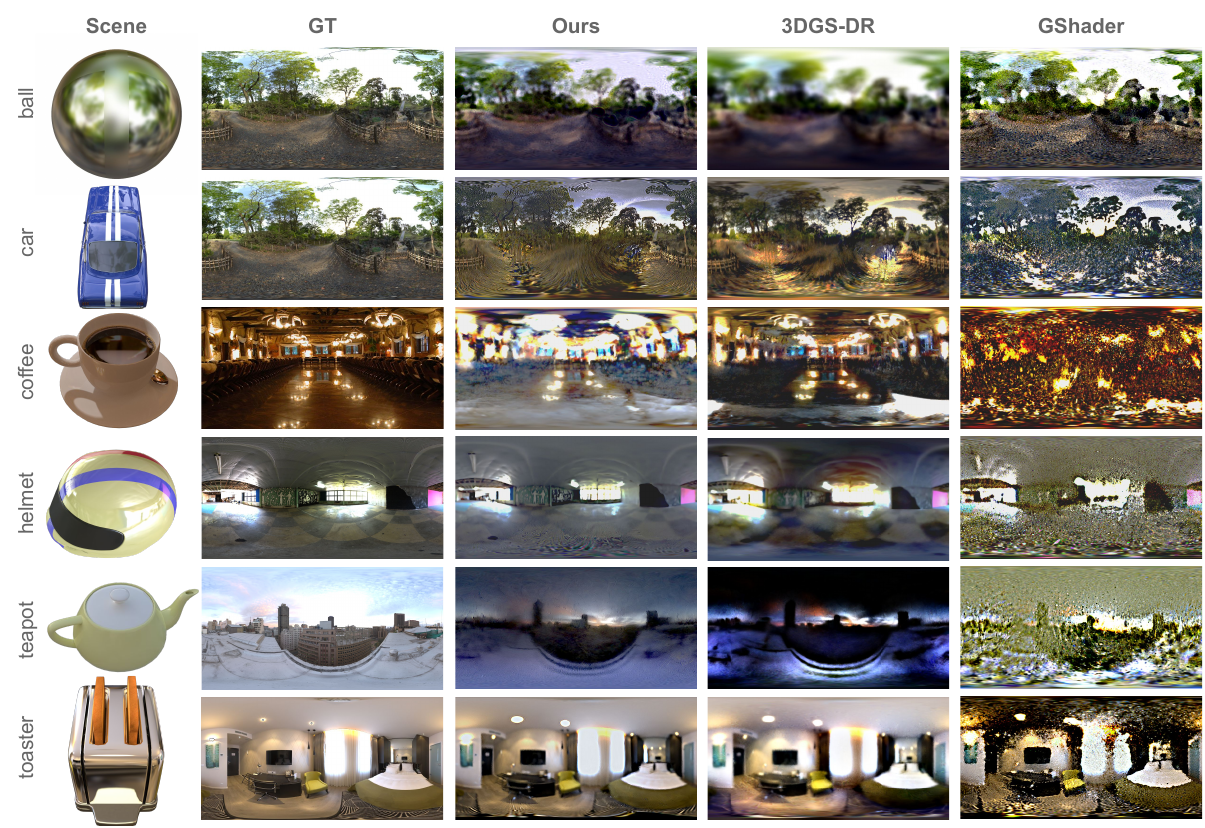}
    \caption{Estimated environment maps on the Shiny Synthetic dataset \cite{verbin2022refnerf} recovered by our RGS-DR, 3DGS-DR~\cite{ye2024gsdr} and GaussianShader~\cite{jiang2024gshader}. In most cases, we recover sharper details with less blurriness or noise than the competing methods.}
    \label{fig:shiny_envmaps}
\end{figure*}

\begin{figure*}[t]
    \centering
    \includegraphics[width=0.82\linewidth,trim=0cm 0cm 0cm 0cm,clip]{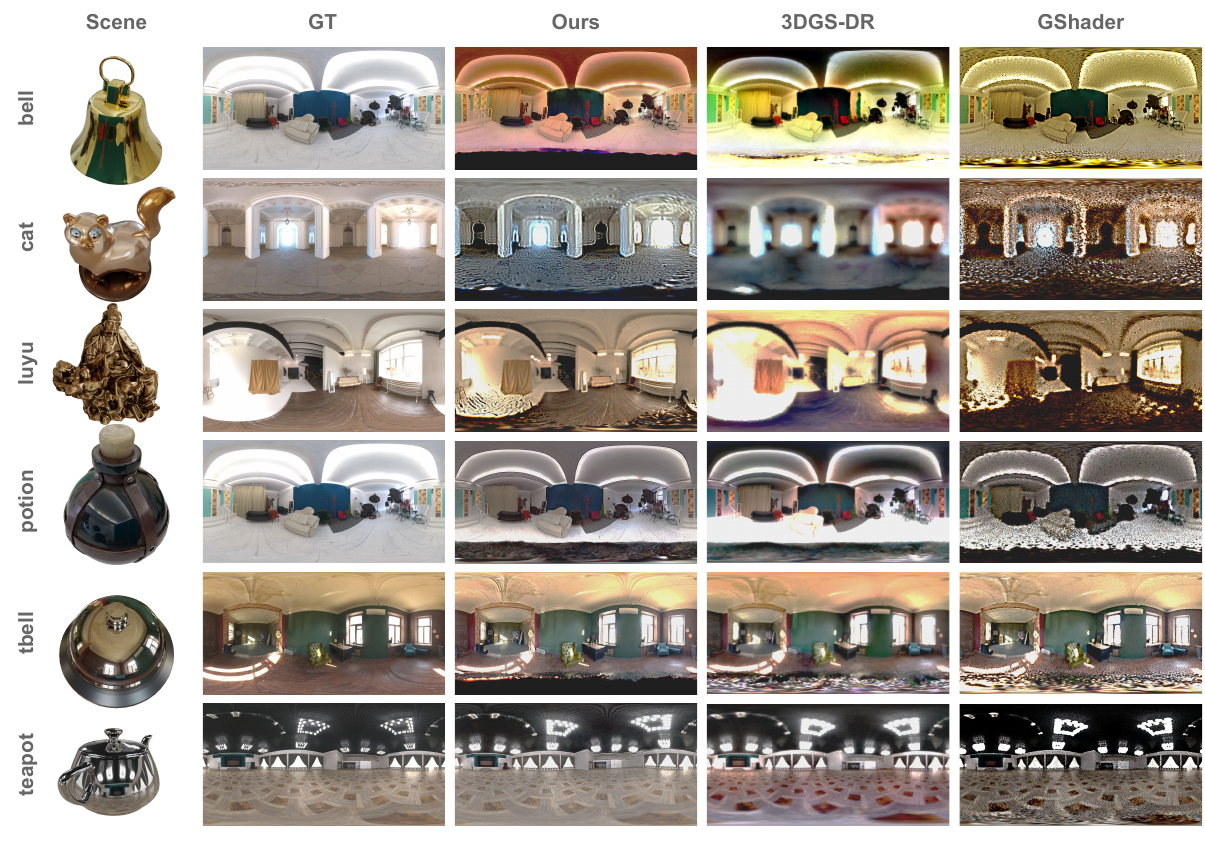}
    \caption{Estimated environment maps on the Glossy Synthetic dataset \cite{liu2023nero} recovered by our RGS-DR, 3DGS-DR~\cite{ye2024gsdr} and GaussianShader~\cite{jiang2024gshader}. In most cases, we recover sharper details with less blurriness or noise than the competing methods.}
    \label{fig:glossy_envmaps}
\end{figure*}

\begin{figure*}[t]
    \centering
    \includegraphics[width=0.8\linewidth,trim=0cm 0cm 0cm 0cm,clip]{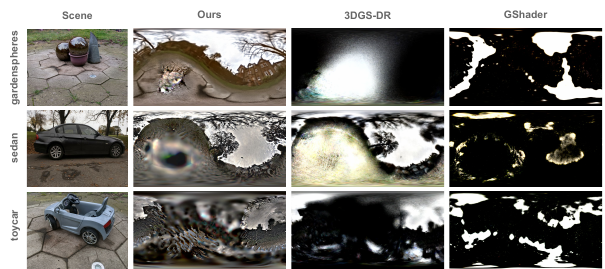}
    \caption{Estimated environment maps on the Shiny Real dataset \cite{liu2023nero} recovered by our RGS-DR, 3DGS-DR~\cite{ye2024gsdr} and GaussianShader~\cite{jiang2024gshader}. In all cases, we recover sharper details with less blurriness or noise than the competing methods.}
    \label{fig:real_envmaps2}
\end{figure*}

In Fig.~\ref{fig:shiny_envmaps}, Fig~\ref{fig:glossy_envmaps}, and Fig.~\ref{fig:real_envmaps2}, we show additional results on estimated environment maps for all evaluated scenes on the Shiny Synthetic \cite{verbin2022refnerf} and Glossy Synthetic~\cite{liu2023nero} datasets. As evident, our method consistently outperforms 3DGS-DR~\cite{ye2024gsdr} and GaussianShader~\cite{jiang2024gshader}, on the shinier scenes e.g. car, helmet, and the entire Glossy Synthetic scenes, while being competitive on the rest. In general, we recover sharper details of the environment illumination with fewer artifacts or blurring. 


\section{Additional Relighting Results}
\begin{figure*}[t]
    \centering
    \includegraphics[width=\linewidth,trim=0cm 0cm 0cm 0cm,clip]{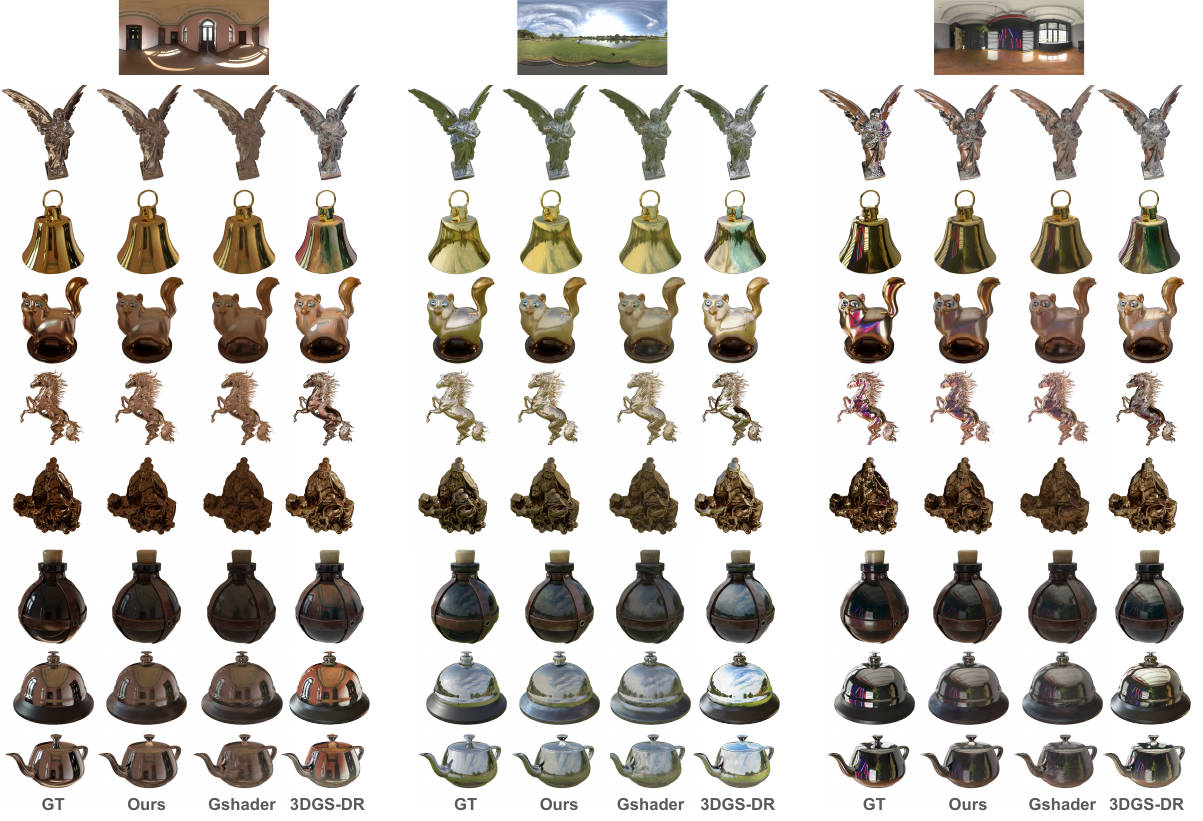}
    \caption{Comparison of relighting results on the Glossy Synthetic scenes \cite{liu2023nero} between our method, GaussianShader~\cite{jiang2024gshader} and 3DGS-DR~\cite{ye2024gsdr}. As can be easily observed, our relighting results are closer to the GT, with sharper reflections and specular highlights.}
    \label{fig:relighting-comparison}
\end{figure*}

\begin{figure*}[t]
    \centering
    \includegraphics[width=0.8\linewidth,trim=0.6cm 0cm 0.5cm 0cm,clip]{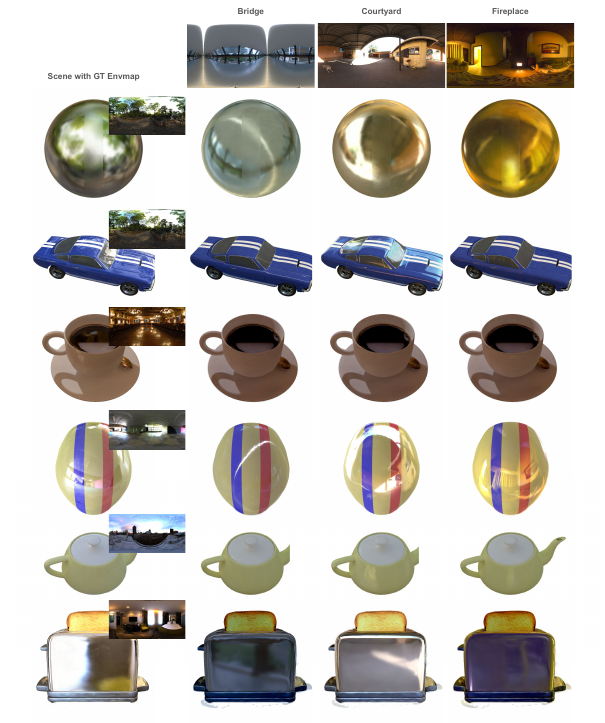}
    \caption{Relighting results on the entire set of scenes of the Shiny Synthetic dataset \cite{verbin2022refnerf}. High-quality and sharp reflections and specular highlights are observed in all scenes and especially the more glossy scenes, showcasing the successful estimate of scene properties. From left to right, we present the optimized scenes with their original environment illumination, followed by three relighting experiments per scene using three distinct environment maps, namely bridge, courtyard, and fireplace.}
    \label{fig:shiny_relightings}
\end{figure*}

\begin{figure*}[t]
    \centering
    \includegraphics[width=0.8\linewidth,trim=0.4cm 0cm 0.3cm 0cm,clip]{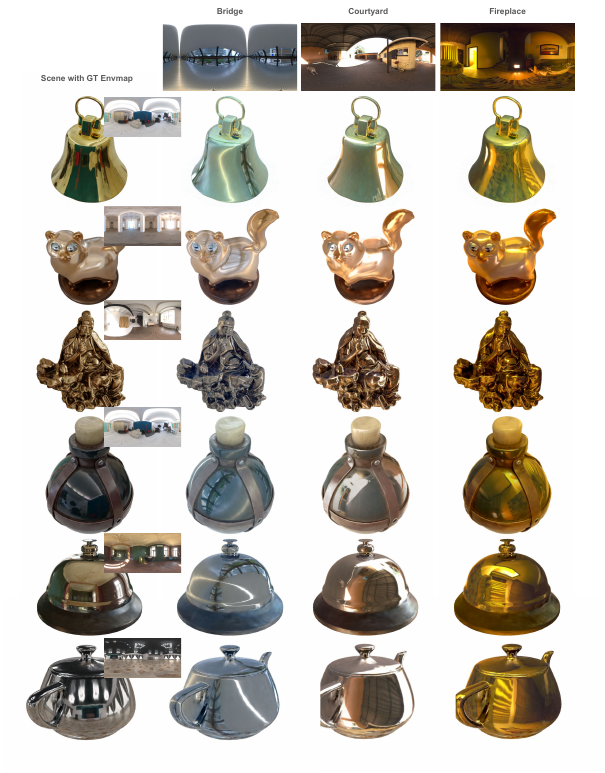}
    \caption{Relighting results on the entire set of examined scenes from the Glossy Synthetic dataset \cite{liu2023nero}. High-quality and sharp reflections and specular highlights are observed in all scenes and especially in the more glossy scenes, showcasing the successful estimate of scene properties. From left to right, we present the optimized scenes with their original environment illumination, followed by three relighting experiments per scene using three distinct environment maps, namely bridge, courtyard, and fireplace.}
    \label{fig:glossy_relightings}
\end{figure*}

In Table~\ref{tab:glossy-relighting} and Fig.~\ref{fig:relighting-comparison}, we present a comparison of our method against GaussianShader \cite{jiang2024gshader} and 3DGS-DR~\cite{ye2024gsdr} on relighting of the Glossy Synthetic \cite{liu2023nero} scenes, demonstrating the superiority of our method both quantitatively and qualitatively. We provide additional qualitative relighting results of our method in Fig.~\ref{fig:shiny_relightings} for the Shiny Synthetic scenes \cite{verbin2022refnerf} and Fig.~\ref{fig:glossy_relightings} for the Glossy Synthetic scenes \cite{liu2023nero} under three distinct environment maps.

\begin{table}[t]
\centering
\setlength{\tabcolsep}{12pt}
\caption{Comparison of relighting performance on the Glossy Synthetic \cite{liu2023nero} dataset.}
\label{tab:glossy-relighting}
\begin{tabular}{l c c c}
\hline
& PSNR $\uparrow$ & SSIM $\uparrow$& LPIPS $\downarrow$ \\
\hline
3DGS-DR & 20.541 & 0.887 & 0.091 \\
GShader & 24.236 & 0.906 & 0.089 \\
Ours & \textbf{24.474} & \textbf{0.920} & \textbf{0.070} \\
\hline
\end{tabular}
\end{table}

\section{Supplementary Video}

In the \href{https://drive.google.com/file/d/1tJBisVljfYMnsNNUS7qHZ4TerOcyqJfM/view?usp=sharing}{supplementary video}, we demonstrate the outputs of our method using smooth animations of reconstructed scenes. First, we showcase the decomposition of a scene into its material properties and light components (diffuse, specular, residual), as well as the underlying geometry (surface normals). Then we show side-by-side comparisons of rendered objects by our method versus 3DGS-DR~\cite{ye2024gsdr} and GaussianShader~\cite{jiang2024gshader}. Our method clearly outperforms both methods on geometry estimation thanks to the utilized 2D Gaussian surfels, showcasing less artifacts, smoother geometry, better appearance reconstruction than GaussianShader and usually on par or better reconstruction than 3DGS-DR. We then present some additional relighting results before concluding the video with more high-quality, rendered animations of glossy objects. 

\section{Limitations}
Our results reveal several limitations. 
First, the deferred shading assumption, that each pixel corresponds to the nearest opaque surface, prevents accurate handling of transparent or refractive materials. This could potentially be addressed by combining deferred rendering for opaque regions with forward rendering for transmissive ones.
Second, the lack of multi-bounce ray tracing and explicit visibility reasoning limits our ability to capture interreflections and shadows, leading to degraded performance on scenes such as coffee and teapot from the Shiny Synthetic dataset \cite{verbin2022refnerf}. Recent methods \cite{chen2025gigs, younes2025texturesplat, yao2025refGS} incorporate indirect light transport into direct IBL, making them better suited for such complex lighting conditions.
Third, our residual pass is only used to improve novel view synthesis and does not contribute to geometry estimation or relighting. Ideally, it could also be leveraged during optimization to compensate for interreflections and other hard-to-model effects, potentially reducing geometric artifacts. Nonetheless, its current role is primarily to narrow the gap with reconstruction-focused methods without explicitly modeling higher-order light transport.  

\end{document}